\documentclass[10pt,twocolumn,letterpaper]{article}

\usepackage{cvpr}              

\usepackage{graphicx}
\usepackage{amsmath}
\usepackage{amssymb}
\usepackage{booktabs}

\usepackage{times}
\usepackage{epsfig}
\usepackage{multirow}
\usepackage[ruled]{algorithm2e}
\usepackage[accsupp]{axessibility}  

\usepackage[pagebackref,breaklinks,colorlinks]{hyperref}

\usepackage[capitalize]{cleveref}
\crefname{section}{Sec.}{Secs.}
\Crefname{section}{Section}{Sections}
\Crefname{table}{Table}{Tables}
\crefname{table}{Tab.}{Tabs.}

\def\cvprPaperID{5455} 
\def\confName{CVPR}
\def\confYear{2022}

\begin{document}

\title{NFormer: Robust Person Re-identification with Neighbor Transformer\vspace{-1mm}}

\author{Haochen Wang$^{1}$, Jiayi Shen$^{1}$, Yongtuo Liu$^1$, Yan Gao$^2$, Efstratios Gavves$^1$\\
University of Amsterdam$^1$, Xiaohongshu Inc$^2$\\
{\tt\small \{h.wang3, j.shen, y.liu6\}@uva.nl, wanjianyi@xiaohongshu.com, egavves@uva.nl}
}

\maketitle

\begin{abstract}
Person re-identification aims to retrieve persons in highly varying settings across different cameras and scenarios, in which robust and discriminative representation learning is crucial. Most research considers learning representations from single images, ignoring any potential interactions between them. However, due to the high intra-identity variations, ignoring such interactions typically leads to outlier features. To tackle this issue, we propose a Neighbor Transformer Network, or NFormer, which explicitly models interactions across all input images, thus suppressing outlier features and leading to more robust representations overall. As modelling interactions between enormous amount of images is a massive task with lots of distractors, NFormer introduces two novel modules, the Landmark Agent Attention, and the Reciprocal Neighbor Softmax. Specifically, the Landmark Agent Attention efficiently models the relation map between images by a low-rank factorization with a few landmarks in feature space. Moreover, the Reciprocal Neighbor Softmax achieves sparse attention to relevant -rather than all- neighbors only, which alleviates interference of irrelevant representations and further relieves the computational burden. In experiments on four large-scale datasets, NFormer achieves a new state-of-the-art. The code is released at \url{https://github.com/haochenheheda/NFormer}.

\end{abstract}

\section{Introduction}



\begin{figure}[t]
\begin{center}
\includegraphics[width=1.0\linewidth]{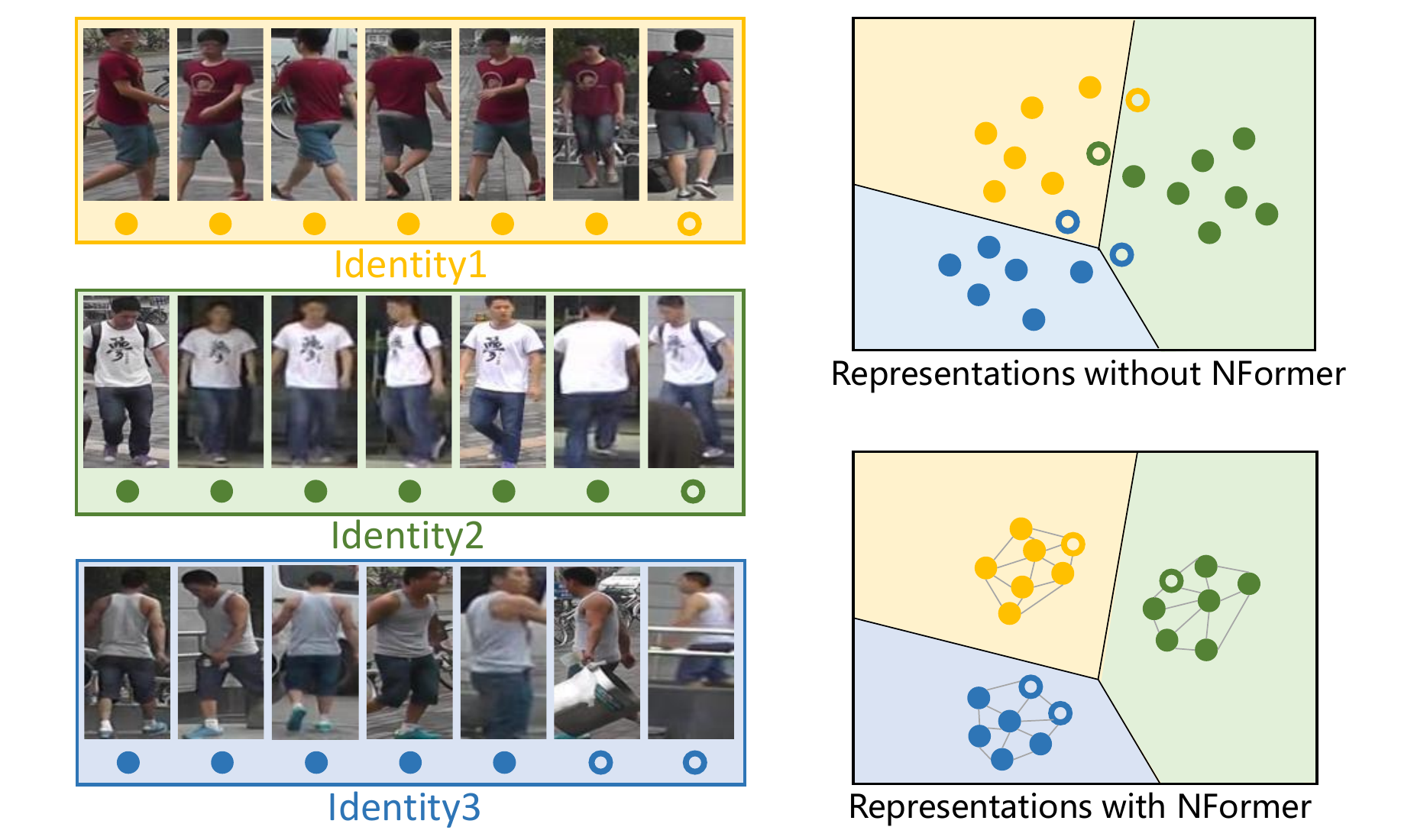}
\end{center}
   \vspace{-7mm}
\caption{The dots indicate the representation vectors of persons for retrieval in the feature space. The right top figure illustrates person representations distribution obtained by learning from single input images, which typically leads to outliers (hollow dots) caused by occlusions, dress changing, viewpoint changing, etc. The right bottom figure shows the person representations distribution obtained by NFormer, which explicitly models the relations (grey lines) between relevant neighbor persons to alleviate the outlier features caused by the above-mentioned abnormal conditions and maintain the most discriminative features for each identity.}
\label{fig:fig1}
\vspace{-4mm}
\end{figure}

Image-based person re-identification (Re-ID) aims to retrieve a specific person from a large number of images captured by different cameras and scenarios. 
Most research to date has focused on how to obtain more discriminative feature representations from single images, either by attention modules \cite{li2018harmonious,shen2018end,wang2018mancs,wang2018person}, part representation learning \cite{suh2018part,zhao2017deeply,cheng2016person,li2017learning}, or GAN generation~\cite{zheng2017unlabeled, liu2018pose,qian2018pose}.
However, one of the main challenges in Re-ID is that any individual typically undergoes significant variations in their appearance due to extrinsic factors, like different camera settings, lighting, viewpoints, occlusions, or intrinsic factors like dress changing, to name a few examples.
As a result, there are high intra-identity variations in the representations corresponding to a specific individual, leading to unstable matching and sensitivity to outliers, see figure \ref{fig:fig1}.

A possible remedy against high intra-identity variations is to exploit the knowledge that exists in the different images from the same identity.
Intuitively, one can encourage the model to cluster neighbor representations tightly, as they are likely to correspond to the same individuals.
A few works have proposed to model relations between input images in Re-ID, either with conditional random fields \cite{chen2018group} or similarity maps from training batches \cite{luo2019spectral}.
However, these works focus on modeling relations between a few images \emph{at training time} only, while during the test, they extract representations per image independently due to the computation limitation, which inevitably loses the interactions and leads to a gap between training and test.
Moreover, they only build relations between a small group of images within each training batches so that there is limited relevant information that could be learned from each other.
To sum up, we argue that encouraging lower representation variations per identity is crucial during both training and test among all the input images.

\begin{figure*}[t]
\begin{center}
\includegraphics[width=0.98\linewidth]{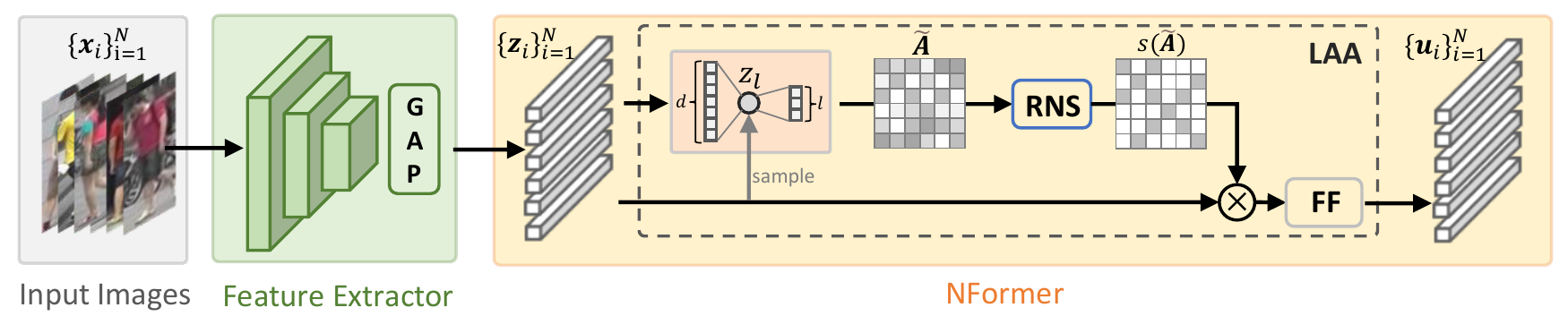}
\end{center}
   \vspace{-5mm}
\caption{An illustration of NFormer. \textbf{GAP}: Global Average Pooling. \textbf{LAA}: Landmark Agent Attention. \textbf{RNS}: Reciprocal Neighbor Softmax. \textbf{FF}: Feed-forward Network. Input with $N$ images $\{\mathbf{x}_i\}_{i=1}^{N}$, a convolutional network followed by GAP is used to get the representation vectors $\{\mathbf{z}_i\}_{i=1}^{N}$. $\{\mathbf{z}_i\}_{i=1}^{N}$ is fed to NFormer, in which LAA is proposed to map the $d$-dimensional representations into a $l$-dimensional encoding space with sampled landmark agents $\mathbf{z}_l$ and then obtain the approximate affinity matrix $\widetilde{\mathbf{A}}$ more efficiently. Then the RNS is proposed to get the sparse attention weights $s(\widetilde{\mathbf{A}})$ and the output representations $\{\mathbf{u}_i\}_{i=1}^N$ are obtained by weighted aggregation of $\{\mathbf{z}_i\}_{i=1}^{N}$. Finally, a ranking algorithm is performed on the representation vectors after NFormer for the retrieval process.}
\label{fig:pipe}
\vspace{-4mm}
\end{figure*}

Following this train of thought, we propose a Neighbor Transformer Network, or NFormer for short, to efficiently model the relations among all the input images \emph{both at training and test time}.
As shown in figure~\ref{fig:pipe}, NFormer computes an affinity matrix representing the relations between the individual representations and then conducts the representation aggregation process according to the affinity matrix.
The involvement of relation modeling between images suppresses high intra-identity variations and leads to more robust features.


Unfortunately, computing an affinity kernel matrix is typical of quadratic complexity to the number of samples.
Such a computational complexity is prohibitively expensive in person Re-ID setting, where the number of input images can easily grow to several thousands during the inference.
To this end, we propose a \emph{Landmark Agent Attention} module (LAA) that reduces the computations in the affinity matrix by the introduction of a handful of landmark agents in the representation space.
The landmark agents map the representation vectors from a high-dimensional feature space into a low-dimensional encoding space, which factorizes large affinity maps into a multiplication of lower rank matrices. 
Similarly, the representation aggregation process with a standard softmax attends to all the input representations, which tends to be distracting and computation-consuming caused by a large number of irrelevant representations.
We introduce the \emph{Reciprocal Neighbor Softmax} function (RNS) to achieve sparse attention attending to computationally manageable neighbors only.
The Reciprocal Neighbor Softmax significantly constrains the noisy interactions between irrelevant individuals, which makes the representation aggregation process more effective and efficient.


Our contributions are summarized as follows:\vspace{-1mm}
\begin{itemize}
\item We propose to explicitly model relations between person representations with a Neighbor Transformer Network, designed to yield robust and discriminative representations.\vspace{-2mm}
\item We design a Landmark Agent Attention module to reduce the computational cost of the large affinity matrix by mapping the representations into lower-dimensional space with a handful of landmark agents.\vspace{-2mm}
\item We propose a Reciprocal Neighbor Softmax function to achieve sparse attention attending to neighbors only, which strengthens the interaction between relevant persons with efficiency.\vspace{-2mm}
\item We conduct extensive experiments on four person Re-ID datasets to indicate the general improvements which NFormer brings. 
The results show that NFormer achieves a new state-of-the-art. 
We further note that NFormer is easy to plug and play with other state-of-the-arts and further boosts the performance.
\end{itemize}

\section{Releated Work}


In this section, we first briefly review two main families of Re-ID methods: Feature Representation Learning methods and Ranking Optimization methods.
Then we introduce the Transformer and related applications.

\subsection{Feature Representation Learning Methods}
\label{related:fpl}
Learning the discriminative feature representations is crucial for Re-ID.
Most of the existing methods \cite{li2018harmonious,shen2018end,suh2018part,zhao2017deeply,zheng2017unlabeled} focus on how to extract better representation with single images. 
Some methods introduce local part features with automatic human part detection \cite{suh2018part, zhao2017deeply} or horizontal image division \cite{sun2018beyond} to tackle the occlusion and misalignment problems.
Some methods design attention modules within single images to enhance representation learning at different levels.
For instance, method \cite{li2018harmonious} involves pixel-level attention while methods \cite{wang2018mancs,wang2018person} achieve channel-wise attention for feature re-allocation.
Method \cite{song2018mask} suppresses the background region to obtain robust foreground person representation.
Another kind of method focuses on increasing the richness of training data.
\cite{huang2018adversarially, zhong2020random} generates adversarially occluded samples to augment the variation of training data.
\cite{zheng2017unlabeled, liu2018pose} utilize GAN to generate images as auxiliary information to help the training.
In general, this family of methods makes full use of information from individual images to extract discriminative feature representations. 

\subsection{Ranking Optimization Methods}
\label{related:ro}
Ranking optimization is a strategy to improve the retrieval performance in the test stage. 
Given an initial ranking list obtained by the distance matrix between query and gallery sets, works \cite{ye2016person,liu2013pop,ma2014query,zhou2017efficient,ye2015ranking} optimize the ranking order by the following methods.
\cite{ye2016person} propose a rank aggregation method by employing similarity and dissimilarity. 
\cite{liu2013pop} involves human feedback to optimize the ranking list. 
Methods \cite{ma2014query, zhou2017efficient} propose the query adaptive retrieval strategy to improve the performance. 
\cite{zhong2017re,ye2015ranking} also utilize contextual information from other images.
Those methods are directly conducted on each initial ranking list as post-processing, instead of conducted on the representation distribution.
Our proposed NFormer is compatible with those re-ranking methods to further boost the performance.

\subsection{Transformer}
\label{related:trans}
The Transformer \cite{vaswani2017attention} is built upon the idea of Multi-Head Self-Attention (MHA), which allows the model to jointly attend to different representation elements.
The transformer is proposed to tackle the sequence problem in the beginning.
Recently, Tranformer is widely used for many vision tasks because of its powerful ability to obtain long-distance dependance, such as DETR \cite{carion2020end} for object detection, TT \cite{chen2021transformer} for object tracking and ViT \cite{dosovitskiy2020image} for image classification.
We first adopt Transformer architecture to learn the relations between input persons in the Re-ID task.

\section{Neighbor Transformer Network}


We first describe the problem setting and the overview of the proposed method.
Then, we describe the Landmark Agent Attention and the Reciprocal Neighbor Softmax.

\subsection{Problem Setting}
\label{nformer:setting}

Re-ID is typically cast as a retrieval task.
We start with the training set $T = {\{\mathbf{x}_i, y_i\}_{i=1}^{N^T}}$, where $\mathbf{x}_i$ corresponds to the $i$-th image with identity $y_i \in S_T$, and $S_T$ contains the identities of all the training images.
During training, we learn a model $\mathbf{z}_i=f(\mathbf{x}_i)$ that computes discriminative feature representations $\mathbf{z}_i$ per input image ~\cite{li2018harmonious,shen2018end,wang2018mancs}.
At test time we have a query set $U = {\{\mathbf{x}_i\}_{i=1}^{N^U}}$ with persons-of-interest.
Then, given a gallery set $G = {\{\mathbf{x}_i\}_{i=1}^{N^G}}$ for retrieval, we retrieve persons with correct identity when comparing query images in $U$ against the images in the gallery set $G$.
The identities of the persons in the query set $S_U$ are disjoint from the identities available during training, that is $S_U \cup S_T = \emptyset$.


\subsection{Learning NFormer}
\label{nformer:overview}

In the described setting, we place no restrictions on the form of the function $f(\cdot)$.
Typically, $f(\cdot)$ is computed on single input images, thus ignoring any possible relations that may arise between the representations of the same individual across cameras and scenarios.
To explicitly account for such relations, we introduce a function to get the aggregated representation vector $\mathbf{u}_i$:\vspace{-1mm}
\begin{align}
\mathbf{u}_i = g(\mathbf{z}_i, \{\mathbf{z}_j\}_{j=1}^{N}) = \sum_{j} \mathbf{w}_{ij} \mathbf{z}_{j},\vspace{-4.0mm}
\label{eq:raw-aggregation}
\end{align}
where $\{\mathbf{z}_j\}_{j=1}^{N}$ contains the representation vectors obtained by the feature extraction function $f(\cdot)$ of all the input images $\{\mathbf{x}_i\}_{i=1}^N$.
During training, $\{\mathbf{x}_i\}_{i=1}^N\subset T$ is a large batch sampled from training set. 
While during test, $\{\mathbf{x}_i\}_{i=1}^N=U\cup G$ contains all the images from query set and gallery set.
$\mathbf{w}_{ij}$ is learnable weight between $\mathbf{z}_i$ and $\mathbf{z}_j$, where 
$\sum_j \mathbf{w}_{ij} = 1$.
Recently, Transformer \cite{vaswani2017attention} has shown to be particularly apt in modeling relations between elements in a set.
With a Transformer formulation, we have equation \eqref{eq:raw-aggregation} reformed by:
\vspace{-0.5mm}
\begin{align}
\mathbf{u}_i = \sum_{j} s(\mathbf{A})_{ij} \varphi_v(\mathbf{z}_{j}),
\label{eq:aggregation}
\end{align}
\vspace{-1mm}
where $\mathbf{A} \in \mathbb{R}^{N \times N}$ is an affinity matrix that contains the similarities between any two pairs of input representation vectors $\mathbf{z}_i, \mathbf{z}_j$,  $s(\cdot)$ is a softmax function to turn the affinities into weights, and
$\varphi_v(\cdot)$ is a linear projection function.
For the affinity matrix $\mathbf{A}$, we have
\begin{align}
\mathbf{A}_{ij} = K(\varphi_q(\mathbf{z}_{i}), \varphi_k(\mathbf{z}_{j}))/\sqrt{d} = \mathbf{q}_{i}^{\top} \mathbf{k}_{j}/\sqrt{d},
\label{equa:equa1}\vspace{-2mm}
\end{align}
where $\varphi_q(\cdot), \varphi_k(\cdot)$ are two linear projections, which map the input representation vectors $\mathbf{z} \in \mathbb{R}^{N \times d}$ to query and key matrices $\mathbf{q}, \mathbf{k} \in \mathbb{R}^{N \times d}$. 
$N$ is the number of input images and $d$ is the dimension of the representation vectors.
The $K(\cdot, \cdot)$ is typically the inner product function.

%



Unfortunately, considering a conventional transformer network to model relations between the representations of persons in Re-ID is computationally prohibitive.
First, computing the affinity matrix $\mathbf{A}$ in equation~\eqref{equa:equa1} has quadratic $O(N^{2}d)$ complexity with respect to the number of images $N$.
Thus, the computation of affinity matrix scales poorly with $N$, especially when the dimension $d$ of the representation vector is also large.
To this end, we introduce the Landmark Agent Attention module to factorize the affinity computation into a multiplication of two lower-dimensional matrices, which relieves the computational burden of the affinity matrix.
Second, in equation~\eqref{eq:aggregation} for the final representation vector $\mathbf{u}_i$ we attend to all the $\mathbf{z}_j, j \in \{1,\dots,N\}$ to compute the weighted aggregation, which also scales poorly with $N$.
Importantly, the weighted aggregation tends to be noisy and dispersed, caused by a large number of mostly irrelevant images. 
To tackle those problems, we introduce the Reciprocal Neighbor Softmax function to achieve sparse attention to neighbors, which reduces the noisy interactions with irrelevant individuals and makes the representation aggregation more effective and efficient.
We illustrate the full pipeline in figure \ref{fig:pipe}.\vspace{-2mm}




\begin{figure}[t]
\begin{center}
\includegraphics[width=1\linewidth]{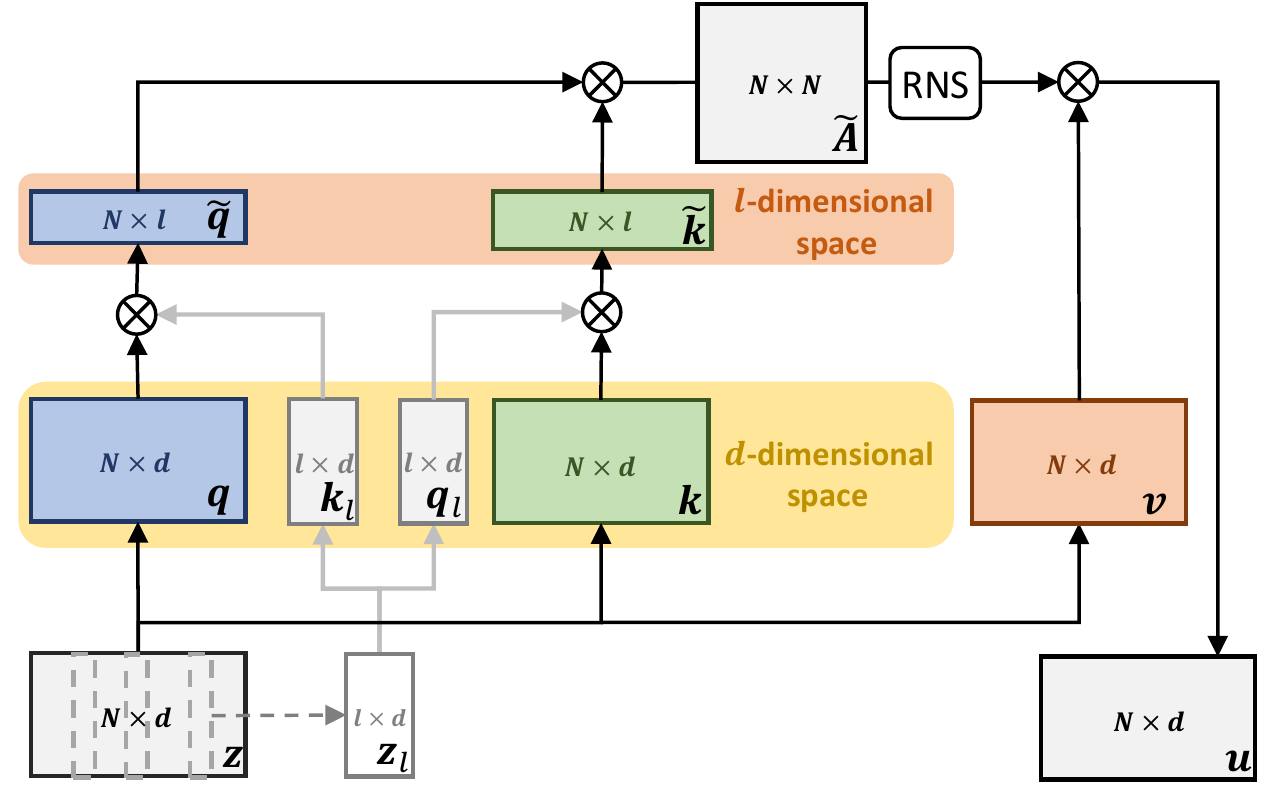}
\end{center}
   \vspace{-5mm}
\caption{Pipeline of LAA. The horizontal side of the rectangles indicates the first dimension of the according matrices, while the vertical side indicates the second dimension. Input with representation vectors $\mathbf{z} \in \mathbb{R}^{N \times d}$, the query, key and value matrices $\mathbf{q}, \mathbf{k}, \mathbf{v} \in \mathbb{R}^{N \times d}$ are generated by three linear projection functions respectively. The landmark agents $\mathbf{z}_l  \in \mathbb{R}^{N \times l}$ are sampled from $\mathbf{z}$ to map the $\mathbf{q}, \mathbf{k}$ of $d$-dimension to $\widetilde{\mathbf{q}}, \widetilde{\mathbf{k}}$ of $l$-dimension. Then the approximate affinity matrix $\widetilde{\mathbf{A}}$ is obtained by the multiplication of $\widetilde{\mathbf{q}}$ and  $\widetilde{\mathbf{k}}$. In this way, the time complexity of obtaining the affinity matrix reduces from $O(N^2d)$ to $O(N^2l)$, since the $l$ is much smaller than $d$ in practice. Then, the RNS is applied to $\widetilde{\mathbf{A}}$ and turns the affinities into sparse attention weights. The final output $\mathbf{u}$ is obtained by weighted aggregation of value matrix $\mathbf{v}$.}
\label{fig:fig2}
\vspace{-3mm}
\end{figure}
\subsection{Landmark Agent Attention}
\label{nformer:laa}

Instead of measuring the similarity between high-dimensional representation vectors, we propose a more efficient way to obtain an approximate affinity matrix $\widetilde{\mathbf{A}}$. 
The key idea is to map the high-dimensional representation vectors $z$ into a lower-dimensional encoding space, making the affinity computations in equation~\eqref{equa:equa1} considerably more efficient, as inspired by random Fourier features~\cite{rahimi2007random}.

As shown in figure \ref{fig:fig2}, following Transformer~\cite{vaswani2017attention}, the query, key and value matrices $\mathbf{q}, \mathbf{k}, \mathbf{v} \in \mathbb{R}^{N \times d}$ are obtained by three separate linear projections $\varphi_q(\cdot), \varphi_k(\cdot), \varphi_v(\cdot)$ using representation vectors $\mathbf{z} \in \mathbb{R}^{N \times d}$ as input. 
Specifically, we randomly sample $l$ representations $\mathbf{z}_l \in \mathbb{R}^{l \times d}$ from $\mathbf{z}$ as landmark agents, and then obtain the $\mathbf{q}_l$ and $\mathbf{k}_l$ matrices with $\varphi_q(\cdot)$ and $\varphi_k(\cdot)$. Thus we could map the original query and key matrices $\mathbf{q},\mathbf{k} \in \mathbb{R}^{N \times d}$ to a $l$-dimensional space by $\widetilde{\mathbf{q}} = \mathbf{q}\mathbf{k}_l^{\top}, \widetilde{\mathbf{k}} = \mathbf{k}\mathbf{q}_l^{\top}$, where $\widetilde{\mathbf{q}}, \widetilde{\mathbf{k}} \in R^{N \times l}$. 
$\widetilde{\mathbf{q}}_{ij}, \widetilde{\mathbf{k}}_{ij}$ indicate the similarity between representation vector $i \in \{1,\dots,N\}$ and landmark agent $j \in \{1,\dots,l\}$.
Then the equation \eqref{equa:equa1} could be replaced by:
\begin{equation}
\begin{aligned}
\widetilde{\mathbf{A}}_{ij} = (\mathbf{q}\mathbf{k}_l^{\top})_{i} (\mathbf{k}\mathbf{q}_l^{\top})^{\top}_j/\sqrt{d} = \widetilde{\mathbf{q}}_{i} \widetilde{\mathbf{k}}^{\top}_{j}/\sqrt{d}.
\end{aligned}
\label{equa:equa3}
\end{equation}
In this way, we decompose the computation of the large affinity map $\mathbf{A} \in \mathbb{R}^{N \times N}$ into a multiplication of two low-rank matrices $\widetilde{\mathbf{q}}, \widetilde{\mathbf{k}}$.
Thus, the multiplication complexity for obtaining the affinity matrix is significantly reduced from $O(N^2d)$ to $O(N^{2}l)$, since $l$ is typically much smaller than $d$ ($l=5,  d\geq 256$ in our experiments). 
In the Supplementary Material section \textcolor[rgb]{1,0,0}{A}, we further prove that the cosine similarity of $\mathbf{A}$ and $\widetilde{\mathbf{A}}$ is positively correlated with $l$, with larger $l$ yielding a cosine similarity close to 1, 
\begin{equation}
\begin{aligned}
\cos( \rm{vec}(\mathbf{A}), \rm{vec}(\widetilde{\mathbf{A}}^{l_b})) \ge \cos(\rm{vec}(\mathbf{A}), \rm{vec}(\widetilde{\mathbf{A}}^{l_a})),
\end{aligned}
\label{equa:prove}\vspace{-1mm}
\end{equation}
where $l_b > l_a$. 
In fact, as we show experimentally in figure \ref{fig:ablation-laa} (a), even with a small number of landmark agents, the NFormer is able to perform stably.

\subsection{Reciprocal Neighbor Softmax}
\label{nformer:rns}

\begin{figure}[t]
\begin{center}
\includegraphics[width=1\linewidth]{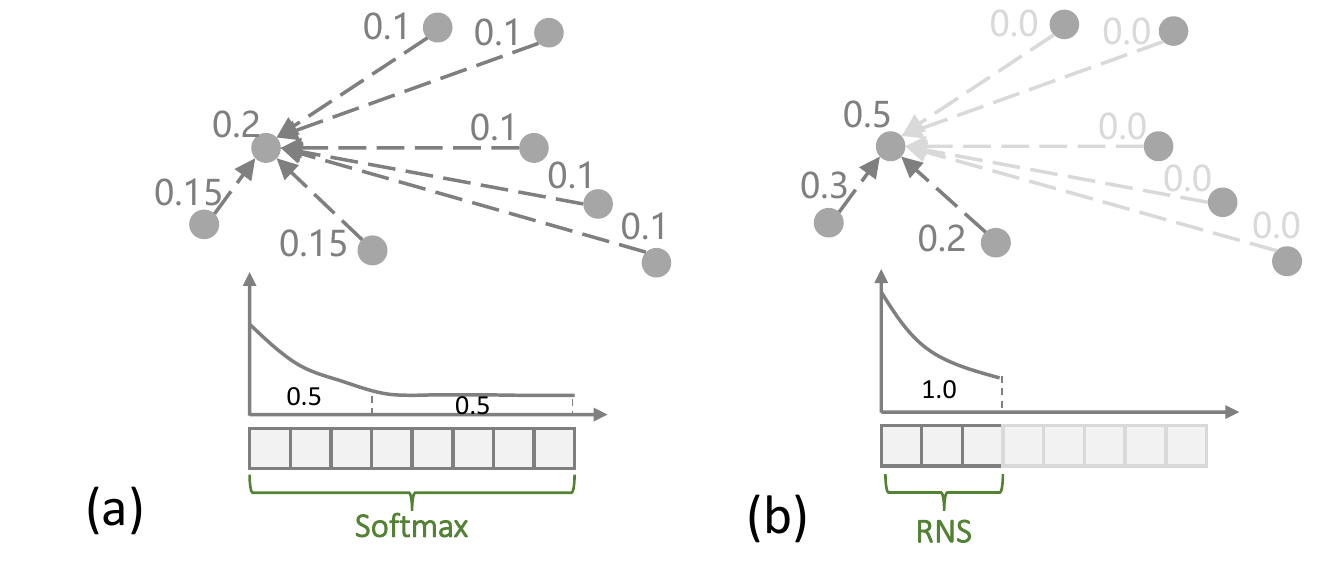}
\end{center}
   \vspace{-5mm}
\caption{Illustration of Reciprocal Neighbor Softmax. (a) indicates the normal softmax, in which the softmax is performed on all the input representations, thus lots of irrelevant representations will contribute to the feature aggregation and distract the attention. (b) indicates the Reciprocal Neighbor Softmax, in which only the relations between reciprocal neighbors are kept.}
\label{fig:fig3}
\vspace{-4mm}
\end{figure}

After obtaining the approximate affinity matrix $\widetilde{\textbf{A}}$, a softmax function $s$ is typically used in equation~\eqref{eq:aggregation} to turn the affinities into attention weights (probabilities).
We can rewrite equation~\eqref{eq:aggregation} as a sum of two parts, $\mathbf{u}_i=\sum_{j: \widetilde{\mathbf{A}}_{ij} \leq \rho} s(\widetilde{\mathbf{A}})_{ij} \varphi_v(\mathbf{z}_j)+\sum_{j: \widetilde{\mathbf{A}}_{ij} > \rho} s(\widetilde{\mathbf{A}})_{ij}\varphi_v(\mathbf{z}_j)$, where $\rho$ is a small threshold. The first part represents the sum of elements with small attention weights and the second part represents the sum of elements with large attention weights.
Although each of the attention weights in $\sum_{j: \widetilde{\mathbf{A}}_{ij} \leq \rho} s(\widetilde{\mathbf{A}})_{ij} \varphi_v(\mathbf{z}_j)$ is small, with a growing number of samples $N$ the total summation will still be large and comparable to the second term in the summation, as shown in figure \ref{fig:fig3} (a).
As a result, the final computation of $\mathbf{u}_i$ will be negatively impacted by the significant presence of irrelevant samples.
Besides the negative effect in the output representation $\mathbf{u}_i$, the computation complexity of the representation aggregation is $O(N^2d)$, which presents a significant computational burden because of the large input size $N$.

To mitigate the above problems, we propose the Reciprocal Neighbor Softmax (RNS) to enforce sparsity to few relevant attention weights with a reciprocal neighbor mask.
We assume that if two images are reciprocal neighbors with each other in feature space, they are likely to be relevant.
To this end, we propose to compute a top-$k$ neighbor mask $\mathbf{M}^k$ from the approximate affinity map $\widetilde{\mathbf{A}}$, which will attend to the top-$k$ value of affinities per row:
\begin{align}
\mathbf{M}^{k}_{ij} = \begin{cases}
1, & j \in {\rm topk}(\widetilde{\mathbf{A}}_{i,:}) \\
0, & \text{otherwise.} \\
\end{cases}
\label{equa:equa2}
\end{align}
We can then obtain a reciprocal neighbor mask $\mathbf{M}$ by multiplying $\mathbf{M}^{k}$ with its transposition using Hadamard Product.
\begin{equation}
\begin{aligned}
\mathbf{M}_{ij} & = \mathbf{M}^{k} \circ \mathbf{M}^{k\top} \\
& = \begin{cases}
1, & j \in {\rm topk} (\widetilde{\mathbf{A}}_{i,:}), i \in {\rm topk}(\widetilde{\mathbf{A}}_{:,j}) \\
0, & \text{otherwise}. \\
\end{cases}
\end{aligned}
\label{equa:equa2}\vspace{-2mm}
\end{equation}
For each element $\mathbf{M}_{ij}$, the value will be set to 1 if $i$ and $j$ are both top-$k$ neighbors of each other, 0 otherwise. 
By adding this mask $\mathbf{M}$ to the regular softmax function, we achieve sparse attention only occurring in neighbors, which increases the focus on more relevant images. 
The formula of RNS is shown as follows:

\begin{equation}
\begin{aligned}
{\rm RNS}(\mathbf{A})_{ij} = \frac{\mathbf{M}_{ij} {\rm exp}(-\widetilde{\mathbf{A}}_{ij})}{\sum\limits_{k}\mathbf{M}_{ik}{\rm exp}(-\widetilde{\mathbf{A}}_{ik})},
\end{aligned}
\label{equa:equa2}\vspace{-2mm}
\end{equation}
Since most attention values are set to zero, as shown in figure \ref{fig:fig3} (b), the relations are constrained to the relevant neighbors, making the aggregation in equation~\eqref{eq:aggregation} more focused and robust. 
Furthermore, as we do not need to conduct addition operation for the representations with zero weights, the time complexity of feature aggregation significantly decreases from $O(N^2d)$ to $O(Nkd)$.


\section{Experiments}
We first describe the datasets, the evaluation protocols, and the implementation details of NFormer. 
Then, we conduct extensive ablation studies to demonstrate the effectiveness and efficiency brought by each of the proposed modules. 
Finally, we compare NFormer with other state-of-the-art methods on four large-scale Re-ID datasets. 

\subsection{Datasets and Evaluation Protocols}

We conduct experiments on four widely used large-scale person Re-ID datasets: Market1501 \cite{zheng2015scalable}, DukeMTMC-reID \cite{ristani2016performance}, MSMT17 \cite{wei2018person} and CUHK03 \cite{li2014deepreid} to validate the effectiveness and efficiency of NFormer. 
All the above-mentioned datasets contain multiple images for each identity collected from different cameras or scenarios.
We follow the standard person Re-ID experimental setups.
We use the standard metric in the literature \cite{zheng2015scalable} for evaluation: the cumulative matching characteristic (CMC) curve and mean Average Precision (mAP). 
CMC shows the top K accuracy by counting the true positives among the top K persons in the ranking list. 
The mAP metric measures the area under the precision-recall curve, which reflects the overall re-identification accuracy among the gallery set rather than only considering the top K persons.

\subsection{Implementation}

We adopt ResNet-50 \cite{he2016deep} pre-trained on ImageNet \cite{deng2009imagenet} as the backbone architecture for our feature extractor.
To preserve spatial information, we change the stride convolutions at the last stage of ResNet with dilated convolutions, which leads to a total downsampling ratio of 16. 
We then apply a fully connected layer after the Resnet-50 backbone to reduce the dimension of the embedding vector from 2048 to 256 for efficiency. 
We stack four LAA modules to build the NFormer.
The number of landmark agents $l$ in the LAA module is set to 5 and the number of neighbors $k$ in RNS is set to 20 for a good trade-off between computational cost and performance according to the experimental results. 
During the inference, the interactions between the different query images are eliminated for fair comparison.

For all experiments, the images are resized to a fixed resolution of $256 \times 128$. 
Random horizontal flipping is utilized as data augmentation during training.
We combine the identity loss\cite{zheng2017person}, center loss \cite{wen2016discriminative} and triplet loss \cite{hermans2017defense} to form the total loss function.
The three loss functions are weighted by 1, 1, 0.0005 respectively. 
We use Stochastic Gradient Descent (SGD) as the optimizer. 
The initial learning rate is set to 3e-4 and momentum is set to 5e-4. 
We train the Resnet-50 and NFormer in turn for 160 epochs. 
The batch size is set to 128 for training the Resnet-50 feature extractor and is set to 2048 for training NFormer. We freeze the parameters of Resnet-50 during the NFormer training iteration to achieve such a large batch size. 
All the experiments are conducted with PyTorch on one GeForce RTX 3090.


\subsection{Ablation Study}
\label{exp:ablation}

We conduct comprehensive ablation studies on Market-1501 and dukeMTMC-reID datasets to analyze the effectiveness of LAA and RNS with different hyper-parameters. 
With Res50 we denote the modified ResNet-50 feature extractor without the NFormer, and use it as the baseline. 

\begin{figure*}[t]
\begin{center}
\includegraphics[width=1\linewidth]{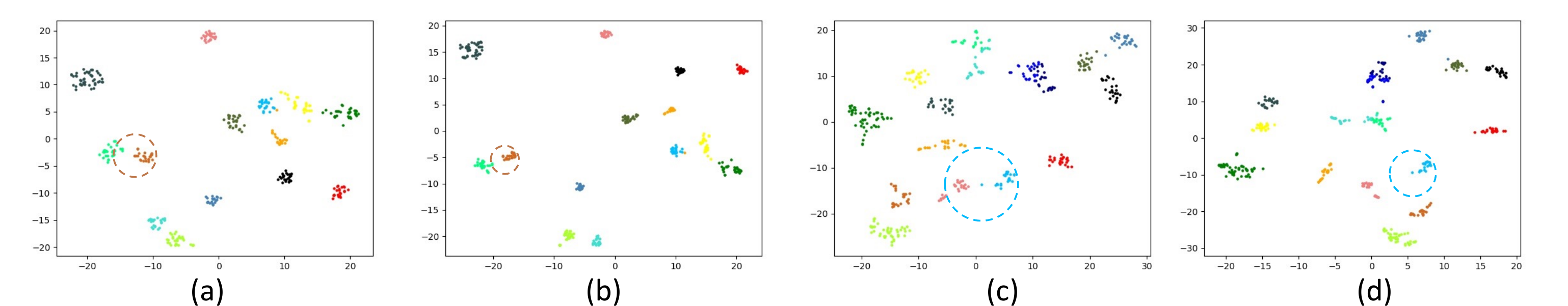}
\end{center}
   \vspace{-5mm}
\caption{t-SNE visualization of representation vectors. (a)/(b) show several random sampled identities on Market-1501 without/with NFormer. (c)/(d) show several random sampled identities on dukeMTMC-reID without/with NFormer. In this figure, we can see that after NFormer, the representation distribution is more gathered and detached. Specifically, if we choose one of the brown points as query person in figure (a), there will be a lot of cyan points at the top of the ranking list, as shown in the brown circle in (a). On the contrary, the ranking list of the same query person contains fewer negative persons in (b) because of the more gathered and detached distribution. The blue circles in (c) and (d) show the same results.}
\label{fig:vis}
\vspace{-3mm}
\end{figure*}

\begin{figure}[h]
\begin{center}
\includegraphics[width=1\linewidth]{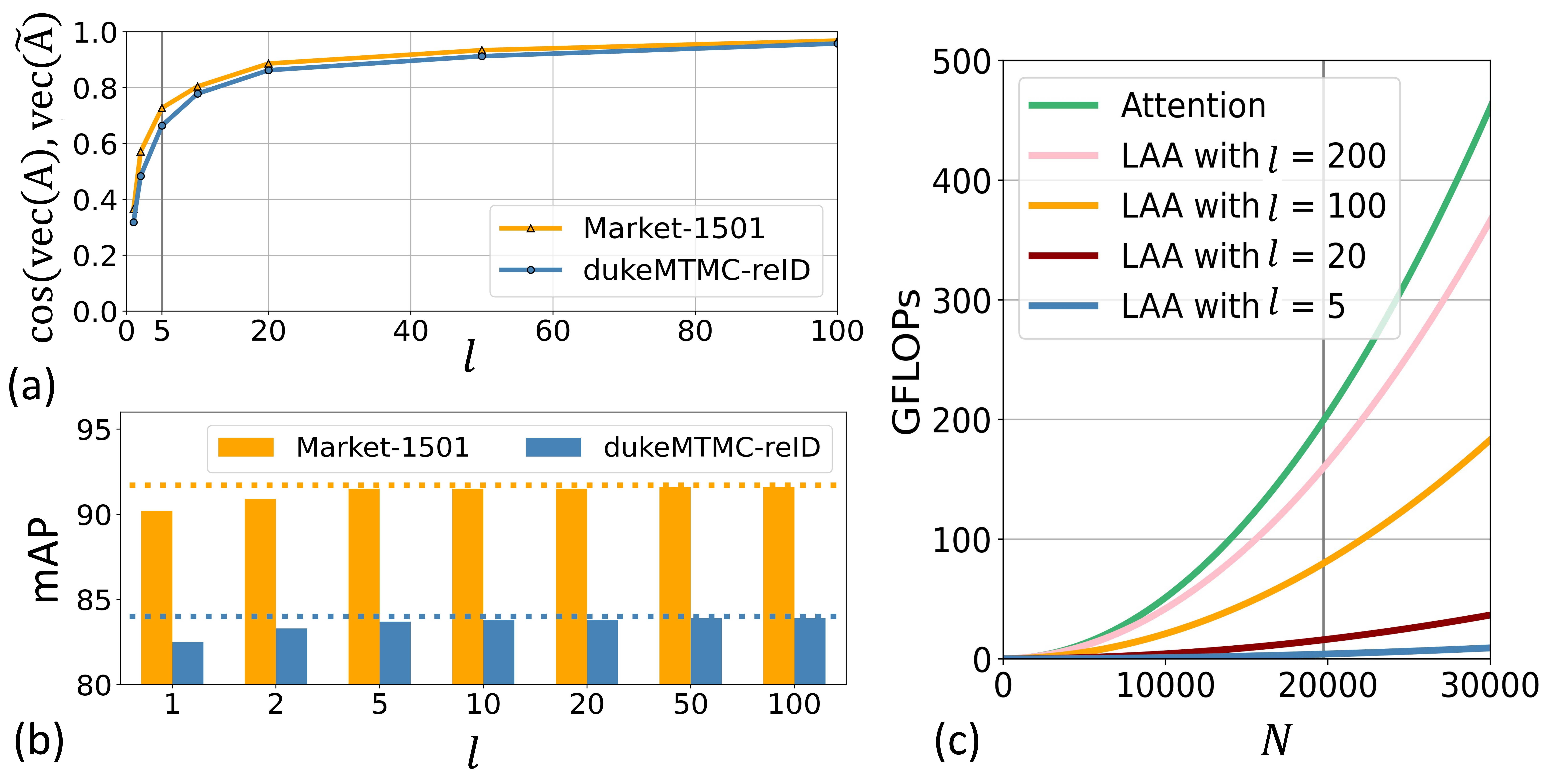}
\end{center}
   \vspace{-6mm}
\caption{Figure (a) shows how the $\cos( \rm{vec}(\mathbf{A}), \rm{vec}(\widetilde{\mathbf{A}}))$ changes with the number of landmark agents $l$. Figure (b) shows how the mAP changes with $l$, in which the orange and blue dash lines show the mAP performance without LAA (with normal affinity matrix). Figure (c) shows how the total GFLOPs changes with input number $N$ under different $l$.} 
\label{fig:ablation-laa}
\vspace{-2mm}
\end{figure}

\begin{figure}[h]
\begin{center}
\includegraphics[width=1\linewidth]{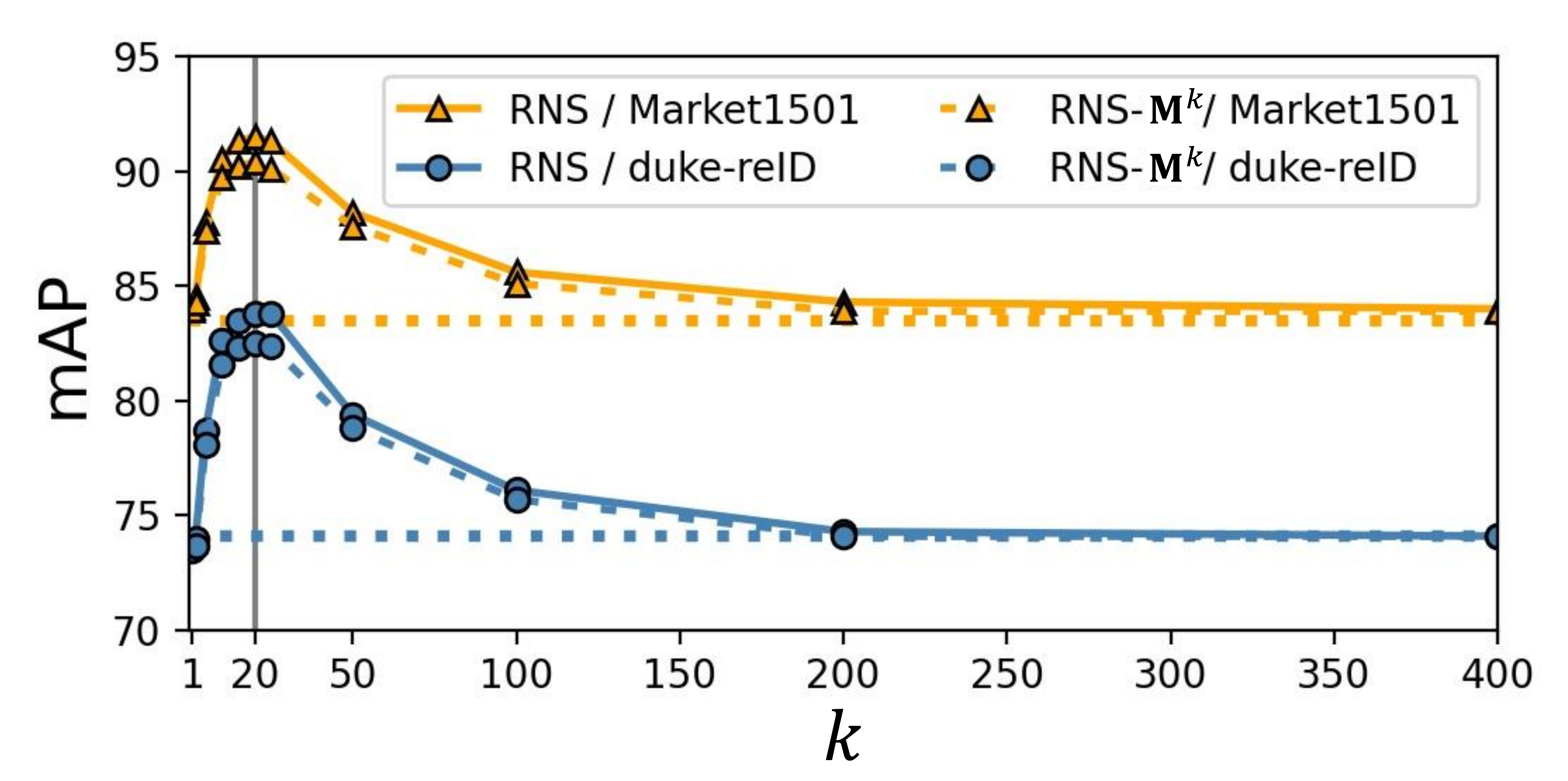}
\end{center}
   \vspace{-7mm}
\caption{This figure shows how mAP changes with the number of neighbors $k$ in RNS on Market-1501 and dukeMTMC-reID. RNS-$\mathbf{M}^k$ indicates RNS with top-k neighbor mask $\mathbf{M}^k$ instead of reciprocal neighbor mask $\mathbf{M}$. The orange and blue horizontal dash lines show the mAP performance with normal softmax function.} 
\label{fig:ablation-rns}
\vspace{-6mm}
\end{figure}

\textbf{NFormer vs. Transformer vs. Res50.}
Table \ref{tab:ablation-base} shows the comparison between NFormer, the regular Transformer and Res50 baseline model on Market-1501 and dukeMTMC-reID datasets. 
The regular Transformer, without any special design, slightly surpasses the baseline model by 0.5\%/1.6\% and 0.5\%/1.3\% top-1/mAP on Market-1501 and dukeMTMC-reID.
With the LAA module and RNS function, NFormer outperforms the baseline model by a considerably larger margin of 1.5\%/7.6\% and 3.3\%/9.4\% top-1/mAP on Market-1501  and dukeMTMC-reID respectively.
Notably, NFormer requires 1.5 orders of magnitude fewer computations (0.0025 GFLOPs vs 0.088 GFLOPs per person for the regular Transformer).

We qualitatively demonstrate the effectiveness of NFormer by visualizing the representation distribution before and after NFormer on Market-1501 and dukeMTMC-reID datasets in figure \ref{fig:vis}.
We observe better feature discriminability for the NFormer, while the outliers of each identity are significantly constrained because the relevant and common information of neighbors is integrated into each data point. 
We conclude that NFormer learns relations between input persons not only effectively but also efficiently.

\begin{table}[t]
\setlength\tabcolsep{3pt}
\centering
\begin{tabular}{l c c c c c}
\toprule
\multirow{2}*{Method} & \multicolumn{2}{c}{Market-1501} & \multicolumn{2}{c}{DukeMTMC} & \multirow{2}*{GFLOPs}\\
 \cmidrule(lr){2-3}
 \cmidrule(lr){4-5}
 & T-1 & mAP & T-1 & mAP &\\ \midrule
Res50 & 93.2&83.5&86.1&74.1 & \\
\midrule
+Transformer\cite{vaswani2017attention} & 93.7&85.1&86.6&75.4 & 0.088\\
+NFormer & 94.7&91.1&89.4&83.5 & 0.0025\\
\bottomrule
\end{tabular}
\vspace{-2mm}
\caption{mAP and GFLOPs comparison between Res50 baseline model, normal Transformer, and NFormer on Market-1501 and dukeMTMC-reID datasets. GFLOPs mean the average number of floating-point operations for processing each input image.\vspace{-4mm}}
\label{tab:ablation-base}
\end{table}

\textbf{Influence of Landmark Agent Attention.}
We first study the effect of the number of landmark agents $l$ for computing the approximate affinity matrix $\widetilde{\mathbf{A}}$. 
Figure \ref{fig:ablation-laa} (a) shows that the cosine similarity $\cos( \rm{vec}(\mathbf{A}), \rm{vec}(\widetilde{\mathbf{A}}))$ is positive relative with the number of landmark agents $l$, and monotonically increasing, approaching 1 even with a small $l$. 
As shown in figure \ref{fig:ablation-laa} (b), the mAP performance on Market-1501 and dukeMTMC-reID achieves 91.1\% and 83.5\% when $l = 5$, which causes only 0.3\% and 0.3\% drops compared with original affinity map without LAA module.
When $l$ gets larger, the cosine similarity and mAP performance are saturated while the FLOPs continue to grow, as shown in figure \ref{fig:ablation-laa} (c).
So we choose $l = 5$ as a good balance between effectiveness and efficiency.
That is, the LAA module only needs 1.95\% of the computations to obtain an approximate affinity map $\widetilde{\mathbf{A}}$, while basically maintaining the performance compared with the original affinity map $\mathbf{A}$.

\textbf{Influence of Reciprocal Neighbor Softmax.}
We show the effect of the number of reciprocal neighbors $k$ in figure \ref{fig:ablation-rns}. When $k$ increases, mAP performance of RNS on Market-1501 and dukeMTMC-reID firstly reaches the maximum values 91.1\% and 83.5\% at $k=20$. 
This is because more neighbors information benefits the aggregation of the individual representations in the early stage.
Then as $k$ continues to increase, the performance gradually decreases because of the introduction of irrelevant interactions.
Therefore, we set $k$ to 20 in all the experiments with RNS.
As shown, RNS outperforms normal Softmax function (horizontal dash lines in figure \ref{fig:ablation-rns}) by 7.3\% and 8.9\% on Market-1501 and dukeMTMC-reID in terms of mAP, which indicates that attending to relevant reciprocal neighbors only leads to better feature representations compared with directly incorporating all the images.
Besides, RNS outperforms RNS-$\mathbf{M}^k$ under the different number of neighbors $k$ consistently, which shows that the reciprocal neighbor mask $\mathbf{M}$ could provide better prior knowledge of learning relations between input images.

\textbf{Complementarity to third methods.}
NFormer is easy to combine with other methods.
We showcase this by choosing a SOTA feature extractor ABD-net~\cite{chen2019abd} for representation learning and a re-ranking method RP~\cite{zhong2017re} to combine with NFormer. 
As shown in table \ref{tab:abl-combine}, NFormer with ABDNet and RP further boosts the performance by 1.0\%/3.0\% and 1.7\%/5.9\% top-1/mAP on Market-1501 and dukeMTMC-reID, which demonstrates the compatibility of NFormer.

\begin{table}[t]
\centering
\begin{tabular}{l c  c c  c}
\toprule
\multirow{2}*{Method} & \multicolumn{2}{c}{Market-1501} & \multicolumn{2}{c}{DukeMTMC}\\
 \cmidrule(lr){2-3}
 \cmidrule(lr){4-5}
 & T-1 & mAP & T-1 & mAP\\ \midrule
Res50 & 93.2&83.5&86.1&76.1\\
+NFormer & 94.7&91.1&89.4&83.5\\
+NFormer+KR\cite{zhong2017re} & 94.6 & 93.0 & 89.5& 88.2\\\midrule
$^*$ABDNet\cite{chen2019abd}  &95.4 & 88.2 &88.7 & 78.6\\
+NFormer &95.7&93.0&90.6&85.7\\
+NFormer+KR\cite{zhong2017re} &95.7 & 94.1& 91.1 & 89.4 \\
\bottomrule
\end{tabular}
\vspace{-2mm}
\caption{Performance of combination of NFormer, ABD-Net and KR re-ranking method on Market-1501 and DukeMTMC datasets. $^*$ represents our reproduced performance. \vspace{-4mm}}
\label{tab:abl-combine}
\end{table}

\textbf{Limitation.} NFormer learns information from neighbor persons in the feature space.
If the number of images for each identity in the testset is small, then the individuals will not be able to obtain a lot of useful information from each other.
We conduct an ablation study on Market-1501 and dukeMTMC-reID datasets to analyze the influence of the average number of images per identity.
Specifically, we sample 4 sub-testsets from the original testsets of Market-1501 and dukeMTMC-reID respectively.
Each sub-testset has a different average number of images per identity.
We then evaluate NFormer and Res50 baseline model on each sub-testset.
The results are shown in table \ref{tab:ablation-number}, from which we can see that as the number of images per identity reduces from 20 to 5, the improvements brought by NFormer $(\Delta {\rm mAP})$ drops significantly from 7.3\%/9.2\% to 2.3\%/3.8\% on Market-1501 and dukeMTMC-reID datasets.
By contrast, the performance of the Res50 baseline model barely changes, and even slightly increases.
The reason is that as the number of images decreases, it is easier to search through the new and smaller test sets.
The results confirm, therefore, that a limitation of the NFormer is that it expects a large enough number of images of the same person.
This makes NFormer particularly interesting in more complex and large-scale settings with many cameras and crowds and less relevant in smaller setups.

\begin{table}[t]
\setlength\tabcolsep{2pt}
\small
\centering
\begin{tabular}{l c c c c c c c c c}
\toprule
\multirow{3}*{Methods} &Dataset& \multicolumn{4}{c}{Market-1501} & \multicolumn{4}{c}{DukeMTMC-reID}\\
 \cmidrule(lr){2-2}
 \cmidrule(lr){3-6}
 \cmidrule(lr){7-10}

& Subset& 0 & 1 & 2 & 3 & 0& 1 & 2 & 3\\ 
 \cmidrule(lr){2-2}
 \cmidrule(lr){3-6}
 \cmidrule(lr){7-10}
 & n/p & 20 & 15 & 10 & 5 & 20& 15 & 10 & 5\\ \midrule
Res50 &\multirow{2}*{mAP}& 83.7 & 83.9 & 84.1 & 85.5& 74.4&74.4&75.2&76.1\\
+NFormer& & 91.0&90.6&90.1&87.8 & 83.6&83.0&81.7&79.9\\\midrule
&$\Delta$mAP&+7.3&+6.7&+6.0&+2.3&+9.2 & +8.6 & +6.5 & +3.8\\
\bottomrule
\end{tabular}
\vspace{-2mm}
\caption{The mAP performance of NFormer and Res50 baseline model on sampled sub-testsets with different n/p of Market-1501 and dukeMTMC-reID datasets. n/p indicates the average number of images per identity. \vspace{-4mm}}
\label{tab:ablation-number}
\end{table}

\begin{table*}[h]
\setlength\tabcolsep{8pt}
\centering
\begin{tabular}{l c c c c c c c c c c}
\toprule
\multirow{2}*{Method} & \multicolumn{2}{c}{Market-1501} & \multicolumn{2}{c}{duke-reID}& \multicolumn{2}{c}{MSMT17}& \multicolumn{2}{c}{CUHK03-L}& \multicolumn{2}{c}{CUHK03-D}\\
 \cmidrule(lr){2-3}
 \cmidrule(lr){4-5}
  \cmidrule(lr){6-7}
   \cmidrule(lr){8-9}
   \cmidrule(lr){10-11}
 & T-1 & mAP & T-1 & mAP& T-1 & mAP& T-1 & mAP& T-1 & mAP\\ \midrule
PCB+RPP (ECCV'18)\cite{sun2018beyond} & 93.8 & 81.6 & 83.3 & 69.2&68.2&40.4&-&-&63.7&57.5\\
GCS (CVPR'18)\cite{chen2018group} &93.5&81.6&84.9 & 69.5&-&-&-&-&-&-\\
MHN (ICCV'19)\cite{chen2019mixed}&95.1&85.0&89.1&77.2&-&-&77.2&72.4&71.7&76.5\\
OSNet (ICCV'19)\cite{zhou2019omni}&94.8&84.9&88.6&73.5&78.7&52.9&-&-&72.3&67.8\\
Pyramid (CVPR'19)\cite{zheng2019pyramidal}&\underline{95.7}&88.2&89.0&79.0&-&-&78.9&76.9&78.9&74.8\\
IANet (CVPR'19)\cite{hou2019interaction} & 94.4 & 83.1 & 87.1 & 73.4&75.5&46.8&-&-&-&-\\
STF (ICCV'19)\cite{luo2019spectral} & 93.4 & 82.7& 86.9 & 73.2&73.6&47.6&68.2&62.4&-&-\\
BAT-net (ICCV'19)\cite{fang2019bilinear}&94.1&85.5&87.7&77.3&79.5&56.8&78.6&76.1&76.2&73.2\\
PISNet (ECCV'20)\cite{zhao2020not}&95.6&87.1&88.8&78.7&-&-&-&-&-&-\\
CBN (ECCV'20)\cite{zhuang2020rethinking}&94.3&83.6&84.8&70.1&-&-&-&-&-&-\\
RGA-SC (CVPR'20)\cite{zhang2020relation}&\textbf{96.1}&88.4&-&-&\underline{80.3}&57.5&\textbf{81.1}&77.4&\underline{79.6}&74.5\\
ISP (ECCV'20)\cite{zhu2020identity}&95.3&88.6&\underline{89.6}&80.0&-&-&76.5&74.1&75.2&71.4\\
CBDB-Net (TCSVT'21)\cite{tan2021incomplete}&94.4&85.0&87.7&74.3&-&-&77.8&76.6&75.4&72.8\\
CDNet (CVPR'21)\cite{li2021combined} &95.1&86.0&88.6&76.8&78.9&54.7&-&-&-&-\\
PAT (CVPR'21)\cite{li2021diverse} &95.4&88.0&88.8&78.2&-&-&-&-&-&-\\
C2F (CVPR'21)\cite{zhang2021coarse} &94.8&87.7&87.4&74.9&-&-&\underline{80.6}&\textbf{79.3}&\textbf{81.3}&\textbf{84.1}\\
\midrule
\midrule
Res50 & 93.2&83.5&86.1&76.1&74.9&50.1&74.7&73.8&73.4&71.2\\
+NFormer & 94.7&\underline{91.1}&89.4&\underline{83.5}&77.3&\underline{59.8}&77.2&78.0&77.3&74.7\\\midrule
$^*$ABDNet(ICCV'19) \cite{chen2019abd} &95.4 & 88.2 &88.7 & 78.6 & 78.4 &55.5&78.7&75.8&77.3&73.2\\
+NFormer &\underline{95.7}&\textbf{93.0}&\textbf{90.6}&\textbf{85.7}&\textbf{80.8}&\textbf{62.2}&\underline{80.6}&\underline{79.1}&79.0&\underline{76.4}\\
\bottomrule
\end{tabular}
\vspace{-2mm}
\caption{Quantitative results on Market-1501, DukeMTMC-reID, MSMT17 and CUHK03 datasets. T-1 means top-1 accuracy and mAP means mean average precision. The best performance value in each column is marked by bold and the second-best performance value is marked by underline. The symbol “-” indicates that the corresponding value is not provided in the corresponding paper. $^*$ represents our reproduced performance.\vspace{-5mm}}
\label{tab:sota}
\end{table*}

\subsection{Comparison with SOTA methods}
\label{exp:sota}

Last, we compare the performance of NFormer with recent state-of-the-art methods on Market1501, DukeMTMC-reID, MSMT17 and CUHK03 in table \ref{tab:sota}. 
Overall, our proposed NFormer outperforms other state-of-the-arts or achieves comparable performance.

\textbf{Results on Market-1501.} As shown in table \ref{tab:sota}, NFormer achieves the best mAP and comparable top-1 accuracy among all the state-of-the-art competitors.
Specifically, even with a simple feature extractor Res50, the mAP of NFormer outperforms the second-best method ISP~\cite{zhu2020identity} (with HRNet-W30\cite{wang2020deep} backbone) by a large margin 2.5\%. 
When combining NFormer with a better feature extractor from ABDNet \cite{chen2019abd}, the mAP/rank-1 accuracy is further boosted by 1.9\%/1.0\% and outperforms the ISP \cite{zhu2020identity} by 4.4\% in terms of mAP. 
Notably, NFormer outperforms methods STF~\cite{luo2019spectral} and GCS~\cite{chen2018group} which build relations inside each training batch by 8.4\% and 9.5\% in terms of mAP.
This indicates that the relation modeling among all input images both during training and test leads to better representations.  
The visualization of the ranking lists is shown in the Supplementary Material section \textcolor[rgb]{1,0,0}{B}, from which we can see that the NFormer could help to constrain the outliers and improve the robustness of the ranking process.

\textbf{Results on DukeMTMC-reID.}  The results are presented in table \ref{tab:sota}, from which we can see that our method outperforms other state-of-the-arts significantly. 
Specifically, NFormer with Res50 feature extractor gains 3.5\% improvement in terms of mAP over second-best method ISP \cite{zhu2020identity}.
NFormer with ABDNet feature extractor outperforms ISP \cite{zhu2020identity} by 1.0\%/5.7\% in terms of top-1/mAP.
We observe that the improvement is more pronounced for the mAP than for the top-1 metric.
The reason is that NFormer reforms the representation of all the input persons, and in general impacts positively the overall search, not just the top retrieval.

\textbf{Results on MSMT17.}  As shown in table \ref{tab:sota}, the NFormer with Res50 feature extractor outperforms the second best method RGA-SC~\cite{zhang2020relation} (ResNet-50 backbone) by 2.3\% in terms of mAP, while NFormer with ABDNet feature extractor outperforms RGA-SC~\cite{zhang2020relation} by 0.5\%/4.7\% in terms of top-1/mAP.
The NFormer outperforms the baseline model significantly by 2.4\%/9.7\% top-1/mAP, which shows that NFormer works even better on larger datasets, as there is rich neighbor information for each person.

\textbf{Results on CUHK03.} We conduct experiments on both the manually labelled version and the detected version of CUHK03 dataset. 
From table \ref{tab:sota}, we can see that the NFormer with ABD-net achieves comparable performance on both labelled and detected sets. 
NFormer with Res50 feature extractor outperforms baseline model by 2.5\%/4.2\% and 3.9\%/3.5\% top-1/mAP on Labelled and Detected sets.
We further illustrate the reasons for the fewer improvements on CUHK03 dataset. 
We count that the average number of images per identity in CUHK03 is 9.6, which is much less than 25.7 in Market-1501, 23.4 in DukeMTMC-reID, and 30.7 in MSMT17. 
So the NFormer can not learn much relevant information from the neighbors.
We provide a detailed analysis in the \textbf{limitation} part in the ablation study.

\section{Conclusion}
In this paper, we propose a novel Neighbor Transformer Network for person re-identification, which interacts between input images to yield robust and discriminative representations.
In contrast to most existing methods focusing on single images or a few images inside a training batch, our proposed method models the relations between all the input images.
Specifically, we propose a Landmark Agent Attention to allow for more efficient modeling of the relations between a large number of inputs, and a Reciprocal Neighbor Softmax to achieve sparse attention to neighbors.
As such, NFormer scales well with large input and is robust to outliers.
In extensive ablation studies, we show that NFormer learns robust, discriminative representations, which are easy to combine with third methods.

{\small
\bibliographystyle{ieee_fullname}
\bibliography{main}
}

\end{document}